\documentclass{article}
\usepackage{spconf,graphicx}
\usepackage{mathtools}
\usepackage{xcolor}
\usepackage{multirow}
\usepackage{pifont}
\usepackage{lmodern}
\usepackage{siunitx}
\usepackage{booktabs}
\usepackage{etoolbox}

\usepackage{svg}
\usepackage{amsmath}
\graphicspath{ {./images/} }

\usepackage[noend]{algpseudocode}
\usepackage{algorithm}

\newrobustcmd{\B}{\bfseries}

\newcommand{\comment}[2][red]{
\ifnum\keepcomment=1
    \ifnum\colorcomment=1
        \textcolor{#1}{\textbf{#2}}
    \else
        \textbf{#2}
    \fi
  \fi
}

\newcommand{\keepcomment}{1}

\newcommand{\colorcomment}{1}
\newcommand\blfootnote[1]{%
  \begingroup
  \renewcommand\thefootnote{}\footnote{#1}%
  \addtocounter{footnote}{-1}%
  \endgroup
}
\DeclareMathOperator{\MemoryRetrieval}{MemoryRetrieval}
\DeclareMathOperator{\MemoryUpdate}{MemoryUpdate}
\DeclareMathOperator{\Proj}{Proj}
\DeclareMathOperator{\Aug}{Aug}
\DeclareMathOperator{\Enc}{Enc}
\DeclareMathOperator{\SGD}{SGD}

\title{Contrastive LEARNING FOR ONLINE SEMI-SUPERVISED GENERAL CONTINUAL LEARNING}
\name{Nicolas Michel, Romain Negrel, Giovanni Chierchia, Jean-François Bercher\thanks{This work has received support from Agence Nationale de la Recherche (ANR) for the project APY, with reference ANR-20-CE38-0011-02. This work was granted access to the HPC resources of IDRIS under the allocation 2022-AD011012603 made by GENCI}}

\address{Univ Gustave Eiffel, CNRS, LIGM, F-77454 Marne-la-Vallée, France}
\begin{document}

\ninept
\maketitle
\begin{abstract}
We study Online Continual Learning with missing labels and propose SemiCon, a new contrastive loss designed for partly labeled data. We demonstrate its efficiency by devising a memory-based method trained on an unlabeled data stream, where every data added to memory is labeled using an oracle. Our approach outperforms existing semi-supervised methods when few labels are available, and obtain similar results to state-of-the-art supervised methods while using only 2.6\% of labels on Split-CIFAR10 and 10\% of labels on Split-CIFAR100.
\end{abstract}

\section{Introduction}
\label{sec:intro}
\blfootnote{Code is available at https://github.com/Nicolas1203/ossgcl}
In the last decade, deep neural networks have demonstrated their efficiency achieving state-of-the-art results in several computer vision tasks, such as image classification or object detection. Efficiently training such networks relies on the following assumptions.
\begin{itemize}
    \item[\bf A1.] Data is identically and independently distributed.
    \item[\bf A2.] Training data can be seen multiple times by the model during the learning process.
    \item[\bf A3.] Training data is fully labeled. 
\end{itemize}
However, these assumptions are seldom true in real-world environments with a continuous data stream. In such scenarios, Assumption A1 cannot be verified, Assumption A2 is difficult to ensure as the amount of data grows indefinitely, and Assumption A3 would imply an oracle labeling every new incoming data. While numerous successful training strategies have been designed to leverage unlabeled data \cite{chen_simple_2020,oord_representation_2019} when A3 cannot be met, coping with the absence of A1 and A2 remains problematic \cite{mai_online_2021}. A known consequence is Catastrophic Forgetting \cite{mccloskey_catastrophic_1989}, and is likely to be observed if no specific measure is taken. Continual Learning (CL) aims to maintain performances in the absence of A1, and Online Continual Learning (OCL) addresses the lack of A1 and A2. 

In this paper, we propose an OCL algorithm for image classification with few labels. Our method is designed to train a neural network in the absence of A1, A2 and A3. We achieve this by introducing a unified Semi-Supervised Contrastive Loss to leverage labeled and unlabeled data. Despite the lack of A3, our results show that the proposed approach is comparable to state-of-the-art supervised OCL methods, and performs better than existing semi-supervised OCL methods when labels are scarce.

The paper is organized as follows. Section \ref{sec:related} discusses the related work. Section 
\ref{sec:OSSGCL} describes our approach: it sets up the problem, introduces a new Semi-Supervised Contrastive Loss, and describes the algorithm. Section \ref{sec:results} assesses the performance of our approach on a standard benchmark. Finally, Section \ref{sec:conclusion} concludes the paper. 

\section{Related Work}
\label{sec:related}
Our method stands at the junction of Contrastive Learning, Continual Learning and Semi-Supervised learning. In the following section we will briefly recall these fields and elaborate onto specific scenarios where label information is incomplete.

\subsection{Contrastive Learning}
Contrastive Learning has become especially popular in recent years for learning representation from data \cite{chen_simple_2020,oord_representation_2019,gutmann_noise-contrastive_2010,wu_unsupervised_2018,chen_improved_2020,khosla_supervised_2020}. The intuition is quite simple. Similar samples (called positives) should have close representations, while dissimilar samples (called negatives) should have their representations as far away as possible.

\textbf{Self-Supervised Contrastive Learning.}  Contrastive Learning was first designed for unlabeled data, where positives are formed by adding  noise to the input. For images, noisy versions are augmentations or views of the original image. All views of the same inputs are positives, while any other image in the training batch is considered a negative. This self-supervision has several drawbacks: large batches are required for sampling enough negatives \cite{chen_simple_2020}, and images from the same class might be considered as negatives \cite{gutmann_noise-contrastive_2010,wu_unsupervised_2018,oord_representation_2019}.

\textbf{Supervised Contrastive Learning.} A supervised contrastive approach has been proposed by Khosla \emph{et al.}\ \cite{khosla_supervised_2020} using label information to overcome some limitations of self-supervision. The authors consider every image from the same class as positives and show significant improvement.


\subsection{Continual Learning}
Continual Learning (CL) has gained popularity in the past years for image classification. The problem is as follows. Consider a sequential learning setup with a sequence $\{\mathcal{T}_1,\hdots,\mathcal{T}_K\}$ of $K$ tasks, and $\mathcal{D}_k=(X_k, Y_k)$ the corresponding data-label pairs. The number of tasks $K$ is unknown and could potentially be infinitely large. For any values $k_1,k_2 \in \{1,\hdots,K\}$, if $k_1\neq k_2$ then we have $Y_{k_1}\cap Y_{k_2}=\emptyset$ and the number of classes in each task is the same. Catastrophic Forgetting occurs when a model's performance drastically drops on past tasks while learning the current task \cite{mccloskey_catastrophic_1989,french_catastrophic_1999}.

\textbf{General Continual Learning.} Early CL studies, and still the majority of current methods, rely on settings where the task-id $k$ is known  \cite{hsu_re-evaluating_2019,kirkpatrick_overcoming_2017,rebuffi_icarl_2016}, or at least the task-boundary is known (i.e. knowing when the task change occurs). However, such information is usually unavailable while training in a real environment \cite{hsu_re-evaluating_2019,lopez-paz_gradient_2017,he_unsupervised_2021,singh_task-agnostic_2021}. Buzzega \emph{et al.} \cite{buzzega_dark_2020} introduced the General Continual Learning (GCL) scenario where task-ids and task-boundaries are unavailable.

\textbf{Online GCL.} Working with a data stream, the model should be able to learn without storing all incoming data. The most realistic scenario Online General Continual Learning (OGCL), described by Buzzega \emph{et al.} \cite{buzzega_dark_2020}, adds one more constraint to GCL: data are presented once in the incoming stream. This means that the model should adapt to current data without having access to all past data, even if we can afford to store a limited amount of stream data. 

\textbf{CL with missing labels.}
While most CL research focuses on the supervised setting, recent works consider CL where few or no labels are available \cite{rao_continual_2019,smith_memory-efficient_2021,boschini_continual_2021}. However, none of them deals with the OGCL setting except the STAM architecture \cite{smith_unsupervised_2021}, where the authors present an online clustering suited for unsupervised OGCL. In this paper we focus on the OGCL with few labels, which we present in Section \ref{sec:OSSGCL}.

\section{Online Semi-Supervised General Continual Learning}
\label{sec:OSSGCL}
In this section we formally define the Online Semi-Supervised General Continual Learning (OSSGCL) setting, and propose an approach to address the underlying problem.

\subsection{Problem Definition}
In the supervised case, each value in $Y=\cup_{k=1}^K Y_k$ is accessible. We consider the semi-supervised case, where we iterate over an incremental unlabeled data stream $\mathcal{S}=\{X_1,\hdots,X_K\}$ and use an oracle to label specifically selected data. In this context, we have access to a subset $Y^l\subset Y$ with $p=\frac{\vert Y^l \vert}{\vert Y\vert}$ being the percentage of labels available. Usually in semi-supervised learning we want $p$ as small as possible. In addition, we consider that we have labeled examples for every classes encountered. We define $X_k^u$ and $(X_k^l, Y_k^l)$ the sets of unlabeled and labeled data for task $k$. The problem then becomes learning sequentially from $\{\mathcal{T}_1,..,\mathcal{T}_K\}$ tasks on partially labeled datasets $\mathcal{D}_k=X_k^u \cup (X_k^l,Y_k^l)$.



\subsection{Proposed Approach}
To tackle the OSSGCL problem, we design a new Semi-supervised Contrastive Loss (SemiCon) combining a Supervised Contrastive Loss \cite{khosla_supervised_2020} and a Self-supervised Contrastive Loss \cite{chen_simple_2020}.

\textbf{Contrastive Learning framework.} Following \cite{chen_simple_2020}, consider the following elements: a data augmentation process $\Aug(\cdot)$ that transforms any given input according to some random procedure with $a \neq b \Leftrightarrow \Aug_a(\cdot) \neq \Aug_b(\cdot)$; an encoder $\Enc_\theta(\cdot)$ that maps the input to the latent space; a projection head $\Proj_\phi(.)$ that maps the latent representation to another space where the loss is applied. The encoder can be any function and the obtained latent representation will be used for downstream tasks. The projection head is discarded when training is complete to keep only the encoder.

\textbf{Multiview batch with missing labels.} 
We define $\mathcal{B}=\{x_i^l, y_i^l\}_{i=1..b_l}\cup \{x_j^u\}_{j=1..b_u}$ as the incoming batch of $b=b_l + b_u$ inputs with labeled and unlabeled data. Considering $a$,$b$ random numbers, we then work on $\mathcal{B}_I = \Aug_a(\mathcal{B})\cup \Aug_b(\mathcal{B})$, the "multiview batch" \cite{khosla_supervised_2020} of $2b$ samples over indices $i \in I$ and $I=I_l\cup I_u$ with $I_l$ the indices over labeled samples and $I_u$ the indexes over unlabeled samples. Moreover, $h_i = \Enc(x_i)$ is the latent representation of $x_i$ with $H_I = \{h_i\}_{i \in I}$, and $z_i = \Proj(h_i)$ is the projection of $h_i$ with $Z_I=\{z_i\}_{i \in I}$. We also define $P(i) = \{j \in I \backslash \{i\}\ \vert\ y_j=y_i\}$ the indices over the positives of $i$ (similar samples), and $\tau$ the temperature.



\textbf{SemiCon Loss.}
We introduce SemiCon, a unified loss designed to train a contrastive model in the context of missing labels. 
We construct this loss by combining two terms. The first one, $\mathcal{L}_\text{m}$ (\ref{eq:Lm}), corresponds to a supervised contrastive loss on labeled memory data with unlabeled streaming data being considered as negatives (dissimilar samples).
\begin{equation}
    \mathcal{L}_\text{m} = \displaystyle{-\sum_{i\in I_l}\frac{1}{\vert P(i) \vert}\sum_{p\in P(i)}\log\frac{e^{z_i\cdot z_p/\tau}}{\displaystyle\sum_{a \in I \backslash \{i\}}e^{z_i\cdot z_a/\tau}}}
    \label{eq:Lm}
\end{equation}
The second term, $\mathcal{L}_\text{u}$ (\ref{eq:Lu}), stands for an unsupervised contrastive loss on unlabeled stream data with labeled steaming data being considered as negatives with $j(i)$ the index such that $i$ and $j(i)$ are indices of augmented samples having the same input source. 
\begin{equation}
    \mathcal{L}_\text{u} = -\displaystyle\sum_{i\in I_u}\log\frac{e^{z_i\cdot z_{j(i)}/\tau}}{\displaystyle\sum_{a \in I \backslash \{i\}}e^{z_i\cdot z_a/\tau}}
    \label{eq:Lu}
\end{equation}
We define a unified loss $\mathcal{L}_{\text{SemiCon}} = \mathcal{L}_\text{m} + \alpha \mathcal{L}_\text{u}$ where $\alpha \in [0,+\infty[$ is a weighting hyper-parameter representing the importance of unlabeled data during training. We can see that $(\forall i \in I_u)\  P(i)=\{j(i)\}$ so the loss can be expressed as
\begin{equation}
    \mathcal{L}_{\text{SemiCon}} = \displaystyle{-\sum_{i\in I}\frac{g_\alpha(i)}{\vert P(i) \vert}\sum_{p\in P(i)}\log\frac{e^{z_i\cdot z_{p}/\tau}}{\displaystyle\sum_{a \in I \backslash \{i\}}e^{z_i\cdot z_a/\tau}}}
    \label{eq:L}
\end{equation}
with $g_\alpha(i)=\alpha$ if $i \in I_l$, and $g_\alpha(i)=1$ otherwise. Methods replaying past data can suffer from overfitting on memory data \cite{chaudhry_tiny_2019} and while SemiCon handles missing labels, it also gives control over how a model should balance learning from memory and streaming data separately. 


\begin{algorithm}[!ht]
    \begin{algorithmic}
    \State\textbf{Input:}\ Unlabeled data stream $\mathcal{S}$; Memory $\mathcal{M}$; $\Aug(.)$; $\Enc_\theta(.)$; $\Proj_\phi(.)$; Oracle $\mathcal{O}$
    \State\textbf{Output:}\ Encoder $\Enc_\theta$; Memory $\mathcal{M}$
    \For{$\mathcal{B}_s \in \mathcal{S}$}\Comment{Unlabeled data}
        \State $\mathcal{B}_m \gets \ \MemoryRetrieval(\mathcal{M})$\Comment{Labeled data}
        \State $\mathcal{B} \gets \mathcal{B}_s \cup \mathcal{B}_m$\Comment{Combined Batch}
        \State $\mathcal{B}_I \gets \Aug_a(\mathcal{B}) \cup \Aug_b(\mathcal{B})$ \Comment{a,b randoms}
        \State $Z_I \gets \Proj_\phi(\Enc_\theta(\mathcal{B}_I))$
        \State $\theta, \phi \gets \SGD(\mathcal{L}_{SemiCon}(Z_I), \theta, \phi)$\Comment{eq.(\ref{eq:L})}
        \State $\mathcal{M} \gets \ \MemoryUpdate(\mathcal{B}_s, \mathcal{O}, \mathcal{M})$
    \EndFor
    \State\textbf{return:}\ $\theta$;\ $\mathcal{M}$
    \end{algorithmic}
\caption{Proposed Training Method\label{alg:proposedmethod}}
\end{algorithm}


\textbf{Training Procedure.} 
We propose an approach inspired by the work of Mai \emph{et al.}\ \cite{mai_supervised_2021}. They defined Supervised Contrastive Replay (SCR) which combines a Supervised Contrastive Loss \cite{khosla_supervised_2020} and a memory-based strategy \cite{rolnick_experience_2019}. Their method achieves state-of-the-art results in online CL when every data is labeled. We adapt SCR to work with a limited amount of labeled data using SemiCon as objective. Our approach relies on two points: (a) each data added to the memory buffer is labeled, (b) we leverage labeled and unlabeled data in a unified contrastive objective using SemiCon.

During the training phase, we iterate over an unlabeled data stream $\mathcal{S}$. For each incoming stream batch $\mathcal{B}_s \in \mathcal{S}$, we randomly sample a labeled data batch $\mathcal{B}_m$ from memory $\mathcal{M}$ and work on $\mathcal{B}=\mathcal{B}_s\cup \mathcal{B}_m$. Each data batch $\mathcal{B}$ is augmented, and the obtained multiview batch $\mathcal{B}_I$ is fed to the network to compute image projections $Z_I$. The objective function $\mathcal{L}_{SemiCon}$ is computed on $Z_I$, and the model parameters are updated using vanilla Steepest Gradient Descent (SGD). After each SGD step, memory data is updated using Reservoir Sampling \cite{vitter_random_1985}. Each selected stream data is labeled using the Oracle $\mathcal{O}$ before memory storage.

During the testing phase, memory data representations $H_\mathcal{M} = \{Enc_\theta(x)\}_{x \in \mathcal{M}}$ are computed and a classifier is trained on $H_\mathcal{M}$. Our approach is detailed in Algorithm \ref{alg:proposedmethod}. Similarly to Mai \emph{et al.}, we use the Nearest Class Mean (NCM) classifier for the testing phase. Any other classifier can be used on top of the obtained representations; however, we found no significant difference in performance.


\section{Experimental Results}
\label{sec:results}
In this section, we describe two CL datasets, introduce baselines and present our results compared to state-of-the-art.

\subsection{Datasets}
We use modified versions of standard image classification datasets \cite{krizhevsky_learning_nodate} to build an incremental learning environment. These datasets are built on CIFAR10 and CIFAR100 by splitting them into several tasks of non-overlapping classes. Specifically, we work on Split-CIFAR10 and Split-CIFAR100. We divide CIFAR10 in 5 tasks with 2 classes per task and CIFAR100 in 10 tasks with 10 classes per task. Each dataset contains 50,000 train images and 10,000 test images.

\subsection{Baselines}
To assess our results, we compare them to several baselines which respect the OGCL setting and are listed below:
\begin{itemize}
    \setlength\itemsep{0pt}
    \setlength{\parskip}{0pt}
    \setlength{\leftskip}{-5mm}
    \item \textbf{offline}: Supervised upper bound. The model is trained without any CL specific constraints.
    \item \textbf{fine-tuned}: Supervised lower bound that trains the model in a CL setting without precautions to avoid forgetting. 
    \item \textbf{SCR}: Current state-of-the-art on fully supervised OGCL and closest method to our work.
    \item \textbf{SCR - Memory Only} (SCR-MO): SCR method, but trained using memory data only.
    \item \textbf{Experience Replay} (ER) \cite{rolnick_experience_2019}: ER is a simple baseline which applies reservoir sampling \cite{vitter_random_1985} for memory update, just as SCR, but trained with a cross entropy loss rather than a contrastive loss.
    \item \textbf{Experience Replay - Memory Only} (ER-MO): ER-MO is essentially ER, but trained using memory data only.
\end{itemize}
 Even though other semi-supervised CL methods exist, none respects the OSSGCL setting and thus cannot be used in the comparison \cite{he_unsupervised_2021,smith_memory-efficient_2021,brahma_hypernetworks_2021}.



\subsection{Implementation Detail}
For every experiment we train a reduced ResNet-18 \cite{he_deep_2015} from scratch following previous works \cite{lopez-paz_gradient_2017,mai_supervised_2021} and the projection layer $Proj_\phi$ is a multi layer perceptron \cite{chen_simple_2020} with one hidden layer, a ReLU activation and an output size of 128. For memory based methods, we use a memory batch size $\vert \mathcal{B}_m \vert$ of 100 on Split-CIFAR10 and of 500 on Split-CIFAR100. More details on the impact of memory batch size can be found in section \ref{sec:results}. For online methods, we use a stream batch size $\vert \mathcal{B}_s \vert$ of 10, which ensures 5,000 SGD steps on both datasets. Each compared method is trained using SGD optimizer with a learning rate of 0.1, no regularization, and a temperature $\tau=0.07$ for contrastive loss. For every experiment we used the same augmentation procedure as in \cite{chen_simple_2020} and for contrastive methods we construct the multiview batch using one augmentation for each view, while the original SCR implementation used the original image as one view and an augmentation as the other view. We obtained experimentally better results using one augmentation for each view. All experiments are performed 10 times. The average results and their standard deviations are shown in the next section. For the offline baseline, we use the same optimizer and network as other methods and train for 50 epochs.

\subsection{Results}
\label{subsec:results}
In the following we briefly recall a standard CL metric, describe the impact of two hyper-parameters and analyze the obtained results.

\textbf{Metrics.}
We use the accuracy averaged across all tasks after training on the last task. This metric is referred to as the final average accuracy \cite{kirkpatrick_overcoming_2017,hsu_re-evaluating_2019}.

\begin{figure}[htbp]
    \centering
    \includegraphics[width=0.85\columnwidth]{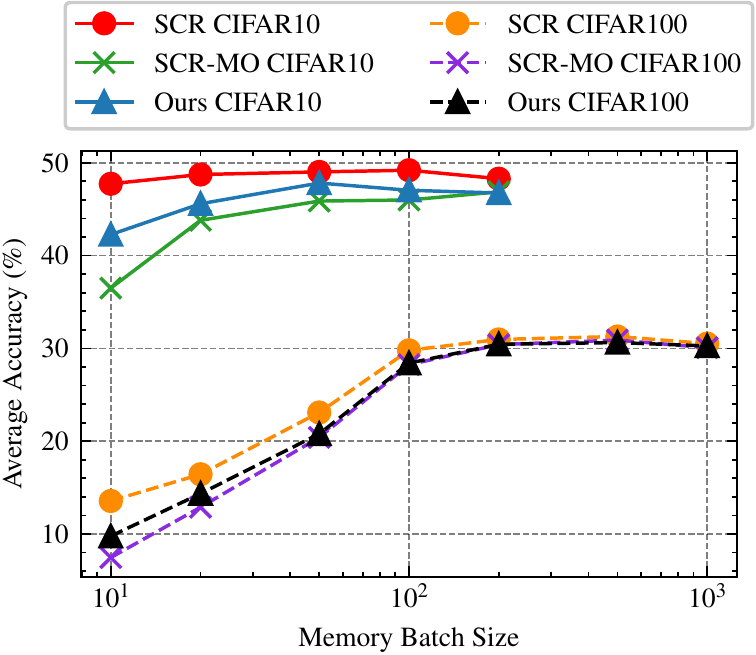}
    \caption{Impact of increasing Memory Batch Size $ \vert \mathcal{B}_m \vert$. We set $\alpha=0$ while augmenting $ \vert \mathcal{B}_m \vert$. The memory size is set to 200 for Split-CIFAR10 and 2k for Split-CIFAR100}
    \label{fig:mem_size}
\end{figure}

\textbf{Memory Batch Size selection.}
We study the impact of the memory batch size $\vert \mathcal{B}_m \vert$ on the performance. We use SCR, SCR-MO and our proposed approach as the case of study. As shown in figure \ref{fig:mem_size}, every method follows the same trend and benefits from larger $\vert \mathcal{B}_m \vert$. We select $\vert \mathcal{B}_m \vert = 100$ for every training on Split-CIFAR10 and likewise $\vert \mathcal{B}_m \vert = 500$ on Split-CIFAR100. Moreover, even with $\alpha=0$, our method consistently outperforms SCR-MO on smaller $\vert \mathcal{B}_m \vert$ values. This demonstrate that using unlabeled negatived can significantly enhance performance when few labels are available.

\begin{figure}[htbp]
    \centering
    \includegraphics[width=\columnwidth]{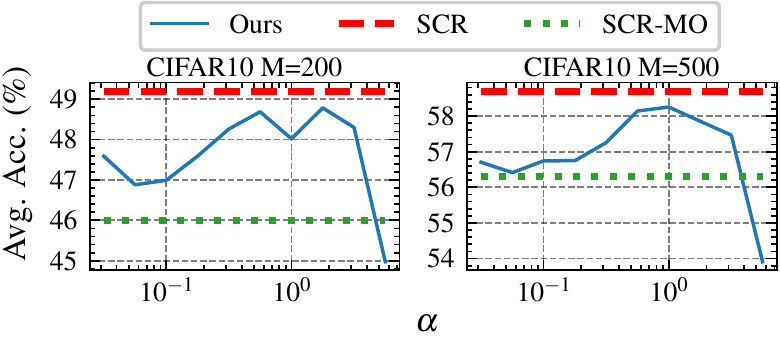}
    \caption{Average Accuracy for $\alpha \in [10^{-1.5},10^{0.75}]$ for our method. Every other parameter is the same for SCR, SCR-MO and our method. Left figure corresponds to Split-CIFAR10 with a memory size M of 200 and right figure corresponds to M=500.}
    \label{fig:alpha}
\end{figure}

\begin{figure}[htbp]
    \centering
    \includegraphics[width=0.9\columnwidth]{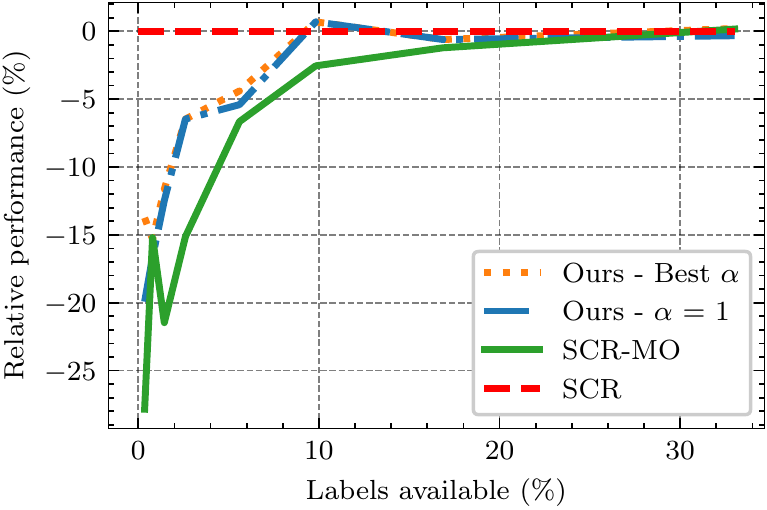}
    \caption{Relative performance compared to SCR on Split-CIFAR100 for SCR-MO and our method while increasing the percentage of available labels.}
    \label{fig:relative}
\end{figure}

\textbf{Impact of $\alpha$.}
We evaluate the impact of $\alpha$ on our method's performance. We keep every other parameter fixed when experimenting on $\alpha$ values. Intuitively, $\alpha$ corresponds to the importance we want to give to unlabeled streaming data against labeled memory data. We observe on figure \ref{fig:alpha} that the optimal value for $\alpha$ depends on the memory size and tends to be close to one. Figure \ref{fig:relative} confirms this observation by comparing performance for $\alpha=1$ to best performance for any $\alpha$ and implies that $\alpha=1$ is an acceptable default parameter for our method. Also, our method performance becomes comparable to SCR-MO for larger memory sizes. We interpret previous observation as the consequence of $\alpha$ playing the role of a regularization parameter. When the memory size is small, the model tends to overfit on memory data and performs better when $\alpha$ is larger. Likewise when the memory size is large, the model has enough information in memory and performs better when $\alpha$ is smaller. Looking at the results on figure \ref{fig:alpha}, best performances are obtained when we use the information from both stream and memory.



\textbf{Results interpretation.} Table \ref{tab:Avg_acc} shows the comparison of our approach to current supervised state-of-the-art methods and their memory-only counterparts on Split-CIFAR10 and Split-CIFAR100 with varying memory sizes M. Our approach achieves best results compared to other semi-supervised methods and performs comparably to supervised state-of-the-art while leveraging only a small portion of labels. This is especially remarkable for small memory sizes where only 2.6\% labels are given to the model. Figure \ref{fig:relative} indicates that our method outperforms SCR-MO when less than 20\% labels are available. Also, our method is on par with SCR using only 10\% labels while SCR-MO needs 20\% labels to obtain comparable results.  Our experiments also show that baselines using the information only available in memory perform competitively, with results close to their supervised counterparts for large memory sizes. This can be explained by the observation that the larger the memory, the closer the problem is to an offline supervised problem.

\begin{table}[ht]
\centering
\resizebox{0.5\textwidth}{!}{
\begin{tabular}{c | c | c c|c c}
    \multicolumn{2}{c|}{Method}                                     & M=200                 & M=500             & M=2k              & M=5k \\
    \hline\hline
    \multirow{4}*{\footnotesize Supervised} & offline               & \multicolumn{2}{c|}{80.0$\pm$1.2}         & \multicolumn{2}{c}{43.2$\pm$2.3} \\
                                            & fine-tuned             & \multicolumn{2}{c|}{16.4$\pm$2.0}  & \multicolumn{2}{c}{3.6$\pm$0.7} \\
                                            & ER                    & 42.6$\pm$1.9          & 52.3$\pm$4.7      & 23.8$\pm$1.3      & 28.5$\pm$1.0 \\
                                            & SCR                   & \B 49.2$\pm$2.2       & \B 58.7$\pm$1.3   & \B 31.3$\pm$0.8   & \B 39.3$\pm$0.8 \\
    \hline
    \multirow{1}*{}                         & \textit{\% labels}    & \textit{2.6\%}        & \textit{5.6\%}    & \textit{16.9\%}    & \textit{33\%} \\
    \multirow{1}*{\footnotesize Semi}       & ER-MO                 & 41.0$\pm$3.5          & 49.9$\pm$3.3      & 23.7$\pm$0.9      & 27.9$\pm$1.0 \\
    \multirow{1}*{\footnotesize Supervised} & SCR-MO                & 46.0$\pm$2.0          & 56.4$\pm$1.4      & 30.9$\pm$0.6      & \B 38.9$\pm$0.8 \\
    \multirow{1}*{}                         & \B Ours               & \B 48.8$\pm$1.1       & \B 57.9$\pm$1.1   & \B 31.0$\pm$0.9   & \B 38.9$\pm$0.5 \\
    
    \hline\hline
    \multicolumn{2}{c}{}                    & \multicolumn{2}{c}{Split-CIFAR10}                   & \multicolumn{2}{c}{Split-CIFAR100}
\end{tabular}}
\caption{Average accuracy on split-CIFAR10 and split-CIFAR100. We use $\alpha=1.78$ and $\alpha=0.18$ for split-CIFAR10 and split-CIFAR100 respectively. Best results for each scenario are displayed in bold. Displayed values correspond to the average and standard deviation over 10 experiments.}
\label{tab:Avg_acc}
\end{table}

\section{Conclusion}
\label{sec:conclusion}
In this paper, we defined a novel OSSGCL setting, and introduced a new Semi-supervised Contrastive Loss (SemiCon). We demonstrated experimentally that semi-supervised approaches trained using memory data only can perform competitively to their supervised counterparts, while leveraging as few as 2.6\% labels on split-CIFAR10. We proposed a new memory-based approach for the OSSGCL setting which successfully combines labeled and unlabeled data using the novel SemiCon loss. This criterion allows user-controlled balance between labeled and unlabeled data during training. We showed that our method can take advantage of unlabeled data, surpassing other semi-supervised baselines on Split-CIFAR datasets, and achieving similar performance to state-of-the-art supervised methods. 

\bibliographystyle{IEEEbib}
\bibliography{refs}

\begin{thebibliography}{10}

\bibitem{chen_simple_2020}
Ting Chen, Simon Kornblith, Mohammad Norouzi, and Geoffrey Hinton,
\newblock ``A simple framework for contrastive learning of visual
  representations,''
\newblock {\em PMLR}, pp. 1597--1607, 2020.

\bibitem{oord_representation_2019}
Aaron van~den Oord, Yazhe Li, and Oriol Vinyals,
\newblock ``Representation learning with contrastive predictive coding,''
\newblock {\em NIPS}, 2018.

\bibitem{mai_online_2021}
Zheda Mai, Ruiwen Li, Jihwan Jeong, David Quispe, Hyunwoo Kim, and Scott
  Sanner,
\newblock ``Online continual learning in image classification: An empirical
  survey,''
\newblock {\em Neurocomputing}, vol. 469, pp. 28--51, 2022.

\bibitem{mccloskey_catastrophic_1989}
Michael McCloskey and Neal~J Cohen,
\newblock ``Catastrophic interference in connectionist networks: The sequential
  learning problem,''
\newblock in {\em Psychology of learning and motivation}, vol.~24, pp.
  109--165. Elsevier, 1989.

\bibitem{gutmann_noise-contrastive_2010}
Michael Gutmann and Aapo Hyv{\"a}rinen,
\newblock ``Noise-contrastive estimation: A new estimation principle for
  unnormalized statistical models,''
\newblock in {\em Proceedings of the thirteenth international conference on
  artificial intelligence and statistics}. JMLR Workshop and Conference
  Proceedings, 2010, pp. 297--304.

\bibitem{wu_unsupervised_2018}
Zhirong Wu, Yuanjun Xiong, Stella~X Yu, and Dahua Lin,
\newblock ``Unsupervised feature learning via non-parametric instance
  discrimination,''
\newblock {\em CVPR}, pp. 3733--3742, 2018.

\bibitem{chen_improved_2020}
Xinlei Chen, Haoqi Fan, Ross Girshick, and Kaiming He,
\newblock ``Improved baselines with momentum contrastive learning,''
\newblock {\em NIPS}, 2018.

\bibitem{khosla_supervised_2020}
Prannay Khosla, Piotr Teterwak, Chen Wang, Aaron Sarna, Yonglong Tian, Phillip
  Isola, Aaron Maschinot, Ce~Liu, and Dilip Krishnan,
\newblock ``Supervised contrastive learning,''
\newblock {\em Advances in Neural Information Processing Systems}, vol. 33, pp.
  18661--18673, 2020.

\bibitem{french_catastrophic_1999}
Robert~M French,
\newblock ``Catastrophic forgetting in connectionist networks,''
\newblock {\em Trends in cognitive sciences}, vol. 3, no. 4, pp. 128--135,
  1999.

\bibitem{hsu_re-evaluating_2019}
Yen-Chang Hsu, Yen-Cheng Liu, Anita Ramasamy, and Zsolt Kira,
\newblock ``Re-evaluating continual learning scenarios: A categorization and
  case for strong baselines,''
\newblock in {\em Continual Learning Workshop of 32nd Conference on Neural
  Information Processing Systems (NeurIPS 2018)}, Montr\'eal, Qu\'ebec, dec
  2018.

\bibitem{kirkpatrick_overcoming_2017}
J.~Kirkpatrick, R.~Pascanu, N.~Rabinowitz, J.~Veness, G.~Desjardins, .~A. Rusu,
  K.~Milan, J.~Quan, T.~Ramalho, A.~Grabska-Barwinska, D.~Hassabis, C.~Clopath,
  D.~Kumaran, and R.~Hadsell,
\newblock ``Overcoming catastrophic forgetting in neural networks,''
\newblock vol. 114, no. 13, pp. 3521--3526,
\newblock Publisher: National Academy of Sciences Section: Biological Sciences.

\bibitem{rebuffi_icarl_2016}
Sylvestre-Alvise Rebuffi, Alexander Kolesnikov, Georg Sperl, and Christoph~H
  Lampert,
\newblock ``{iCaRL}: Incremental classifier and representation learning,''
\newblock pp. 2001--2010, 2017.

\bibitem{lopez-paz_gradient_2017}
David Lopez-Paz and Marc'Aurelio Ranzato,
\newblock ``Gradient episodic memory for continual learning,''
\newblock {\em Advances in neural information processing systems}, vol. 30,
  2017.

\bibitem{he_unsupervised_2021}
Jiangpeng He and Fengqing Zhu,
\newblock ``Unsupervised continual learning via pseudo labels,''
\newblock {\em arXiv:2104.07164}, 2021,
\newblock version: 2.

\bibitem{singh_task-agnostic_2021}
Pranshu~Ranjan Singh, Saisubramaniam Gopalakrishnan, Qiao ZhongZheng,
  Ponnuthurai~N. Suganthan, Savitha Ramasamy, and ArulMurugan Ambikapathi,
\newblock ``Task-{Agnostic} {Continual} {Learning} {Using} {Base}-{Child}
  {Classifiers},''
\newblock in {\em 2021 {IEEE} {International} {Conference} on {Image}
  {Processing} ({ICIP})}, Sept. 2021, pp. 794--798,
\newblock ISSN: 2381-8549.

\bibitem{buzzega_dark_2020}
Pietro Buzzega, Matteo Boschini, Angelo Porrello, Davide Abati, and Simone
  Calderara,
\newblock ``Dark experience for general continual learning: a strong, simple
  baseline,''
\newblock {\em Advances in neural information processing systems}, vol. 33, pp.
  15920--15930, 2020.

\bibitem{rao_continual_2019}
Dushyant Rao, Francesco Visin, Andrei Rusu, Razvan Pascanu, Yee~Whye Teh, and
  Raia Hadsell,
\newblock ``Continual unsupervised representation learning,''
\newblock {\em Advances in Neural Information Processing Systems}, vol. 32,
  2019.

\bibitem{smith_memory-efficient_2021}
James Smith, Jonathan Balloch, Yen-Chang Hsu, and Zsolt Kira,
\newblock ``Memory-efficient semi-supervised continual learning: The world is
  its own replay buffer,''
\newblock {\em IJCNN}, pp. 1--8, 2021.

\bibitem{boschini_continual_2021}
Matteo Boschini, Pietro Buzzega, Lorenzo Bonicelli, Angelo Porrello, and Simone
  Calderara,
\newblock ``Continual semi-supervised learning through contrastive
  interpolation consistency,''
\newblock {\em arXiv preprint arXiv:2108.06552}, 2021.

\bibitem{smith_unsupervised_2021}
James Smith, Cameron Taylor, Seth Baer, and Constantine Dovrolis,
\newblock ``Unsupervised progressive learning and the stam architecture,''
\newblock {\em International Joint Conferences on Artificial Intelligence
  Organization}, 2021.

\bibitem{chaudhry_tiny_2019}
Arslan Chaudhry, Marcus Rohrbach, Mohamed Elhoseiny, Thalaiyasingam Ajanthan,
  Puneet~K. Dokania, Philip H.~S. Torr, and Marc'Aurelio Ranzato,
\newblock ``On {Tiny} {Episodic} {Memories} in {Continual} {Learning},''
\newblock {\em arXiv preprint arXiv:1902.10486}, June 2019.

\bibitem{mai_supervised_2021}
Zheda Mai, Ruiwen Li, Hyunwoo Kim, and Scott Sanner,
\newblock ``Supervised contrastive replay: Revisiting the nearest class mean
  classifier in online class-incremental continual learning,''
\newblock {\em CVPR}, pp. 3589--3599, 2021.

\bibitem{rolnick_experience_2019}
David Rolnick, Arun Ahuja, Jonathan Schwarz, Timothy Lillicrap, and Gregory
  Wayne,
\newblock ``Experience replay for continual learning,''
\newblock in {\em Advances in Neural Information Processing Systems}. vol.~32,
  Curran Associates, Inc.

\bibitem{vitter_random_1985}
Jeffrey~S Vitter,
\newblock ``Random sampling with a reservoir,''
\newblock {\em ACM Transactions on Mathematical Software (TOMS)}, vol. 11, no.
  1, pp. 37--57, 1985.

\bibitem{krizhevsky_learning_nodate}
Alex Krizhevsky,
\newblock ``Learning multiple layers of features from tiny images,''
\newblock M.S. thesis, Department of Computer Science, University of Toronto,
  2009.

\bibitem{brahma_hypernetworks_2021}
Dhanajit Brahma, Vinay~Kumar Verma, and Piyush Rai,
\newblock ``Hypernetworks for continual semi-supervised learning,''
\newblock {\em arXiv preprint arXiv:2110.01856}, 2021.

\bibitem{he_deep_2015}
Kaiming He, Xiangyu Zhang, Shaoqing Ren, and Jian Sun,
\newblock ``Deep residual learning for image recognition,''
\newblock {\em CVPR}, pp. 770--778, 2016.

\end{thebibliography}

\end{document}